%% file: main.tex
\documentclass[letterpaper, 10 pt, conference]{ieeeconf}  %

\IEEEoverridecommandlockouts
\usepackage{cite}
\usepackage{amsmath,amssymb,amsfonts}
\usepackage{algorithmic}
\usepackage{graphicx}
\usepackage{textcomp}
\usepackage{xcolor}
\usepackage{balance}
\usepackage{comment}
\usepackage[ruled,vlined]{algorithm2e}
\usepackage{subcaption}
\usepackage{tensor}
\newcommand{\eg}{\emph{e.g.},}
\newcommand{\ie}{\emph{i.e.},}

\usepackage{hyperref}
\usepackage{float}
\usepackage{rotating}

\def\figref#1{Fig.~\ref{#1}}

\def\eqref#1{Eq.~(\ref{#1})}

\DeclareMathOperator*{\argmax}{arg\,max}

\def\BibTeX{{\rm B\kern-.05em{\sc i\kern-.025em b}\kern-.08em
    T\kern-.1667em\lower.7ex\hbox{E}\kern-.125emX}}
\begin{document}

\title{\LARGE \bf Spectral Geometric Verification: Re-Ranking Point Cloud Retrieval for Metric Localization }
\author{Kavisha Vidanapathirana$^{1,2}$, Peyman Moghadam$^{1,2}$,  Sridha Sridharan$^{2}$, Clinton Fookes$^{2}$   
\thanks{
$^1$ Robotics and Autonomous Systems Group, DATA61, CSIRO, 
Australia. 
E-mails: {\tt\footnotesize \emph{
firstname.lastname
}@data61.csiro.au}}
\thanks{
$^{2}$ Signal Processing, AI and Vision Technologies (SAIVT),
Queensland University of Technology (QUT), Brisbane, Australia.
E-mails: {\tt\footnotesize \emph\{s.sridharan, c.fookes\}@qut.edu.au}
}
}

\maketitle

\input{chapters/abstract}

\input{chapters/introduction.tex}

\input{chapters/related_work.tex}

\input{chapters/method.tex}

\input{chapters/experiments.tex}

\input{chapters/results.tex}

\input{chapters/conclusion.tex}

\bibliographystyle{./bibliography/IEEEtran}
\bibliography{ref}

\end{document}

%% file: chapters/abstract.tex
\begin{abstract}

In large-scale metric localization, an incorrect result during retrieval will lead to an incorrect pose estimate or loop closure. Re-ranking methods propose to take into account all the top retrieval candidates and re-order them to increase the likelihood of the top candidate being correct. However, state-of-the-art re-ranking methods
are inefficient when re-ranking many potential candidates due to their need for resource intensive point cloud registration between the query and each candidate. In this work, we propose an efficient spectral method for geometric verification (named SpectralGV) that does not require registration. We demonstrate how the optimal inter-cluster score of the correspondence compatibility graph of two point clouds represents a robust fitness score measuring their spatial consistency. This score takes into account the subtle geometric differences between structurally similar point  clouds and therefore can be used to identify the correct candidate among potential matches retrieved by global similarity search. SpectralGV is deterministic, robust to outlier correspondences, 
and can be computed in parallel for all potential candidates.
We conduct extensive experiments on 5 large-scale datasets to demonstrate that SpectralGV outperforms other state-of-the-art re-ranking methods and show that it consistently improves the recall and pose estimation of 3 state-of-the-art metric localization architectures while having a negligible effect on their runtime. 
The open-source implementation and trained models are available at: \href{https://github.com/csiro-robotics/SpectralGV}{https://github.com/csiro-robotics/SpectralGV}.

\end{abstract}

%% file: chapters/introduction.tex
\section{Introduction}
\label{sec:intro}

Accurate 6 Degrees of Freedom (DoF) metric localization is an essential component in many applications in embodied intelligence.
For an embodied agent (mobile robot, autonomous car etc.) to autonomously operate in any environment, it must first be able to estimate its pose. 
The task of robust global localization in large-scale environments remains an open problem as current methods struggle to differentiate between structurally similar places with subtle differences and fail to handle degenerate scenes which lack distinct geometric structure. 
When scaling this problem to city-scale environments (or larger), this task is made efficient through a hierarchical two-step formulation. 
This includes the retrieval-based place recognition step, which estimates a set of coarse place candidates, and a pose estimation step which estimates the alignment between the query and the top place candidate to obtain the 6DoF pose.

In the point cloud domain, the tasks of place recognition and 6DoF alignment (\ie{} registration) have been explored largely in isolation\cite{yin2022general, pomerleau2015review, park2019robust}. Recently, several methods have combined the two tasks to offer a complete solution for metric localization \cite{du2020dh3d,cattaneo2022lcdnet,egonn}. They follow a correspondence-based approach, where local-features are used to estimate correspondences between the query point cloud and the top-candidate returned by global-descriptor matching, and the pose is obtained using the estimated correspondences. 
The top place candidate retrieved from place recognition via global descriptor matching may not always be correct due to the challenges outlined above, and an incorrect place candidate will result in an infeasible pose estimation task.

To address such limitations in retrieval, re-ranking methods take into consideration many top retrieval candidates and re-order them such that correct candidates will be ranked higher. Re-ranking is performed using pre-defined criteria specific to each re-ranking method and often involves utilizing additional information complementary to the global descriptors. 
~\figref{fig:bl_pipeline} shows an example scenario where an incorrect top-1 candidate can be replaced with the correct candidate using a re-ranking method.
Re-ranking has become popular in other domains of information retrieval\cite{radenovic2018cnnretrieval, tan2021rerankingtransformer} but has limited utility in point cloud retrieval due to current re-ranking methods either lacking robustness or being inefficient in handling large-scale outdoor point clouds. %
While re-ranking methods used in other domains can be adopted, these methods are not suited to efficiently utilize the precise 3D geometry information present in point clouds.

In this paper, we first demonstrate that geometric-verification is necessary to ensure robust point cloud re-ranking, but these methods are inefficient due their reliance on resource intensive point cloud registration. 
To address this, we introduce an efficient spectral method for geometric verification based re-ranking, named SpectralGV. 
Using the correspondence compatibility graph introduced in Spectral Matching \cite{spectralmatch}, we demonstrate how the optimal inter-cluster score of this graph represents a robust confidence score on the spatial consistency of two point clouds, and utilize it to formulate registration-free geometric verification. 

\input{chapters/tables/main_pipeline.tex}

SpectralGV allows integration with any architecture that extracts both local and global features for a given point cloud, without modifying the architectures and without further training of learning-based methods. It is an architecture agnostic method and shows no bias towards datasets allowing it to generalize well. We integrate our method with 3 state-of-the-art architectures and demonstrate improvements in recall and pose estimation across 5 large-scale datasets in all evaluation settings. SpectralGV outperforms other re-ranking methods while being the only geometric verification method with sub-linear time complexity, enabling real-time deployment.

%% file: chapters/tables/main_pipeline.tex
\begin{figure*}[t]
    \centering
    \includegraphics[width=0.99\textwidth]{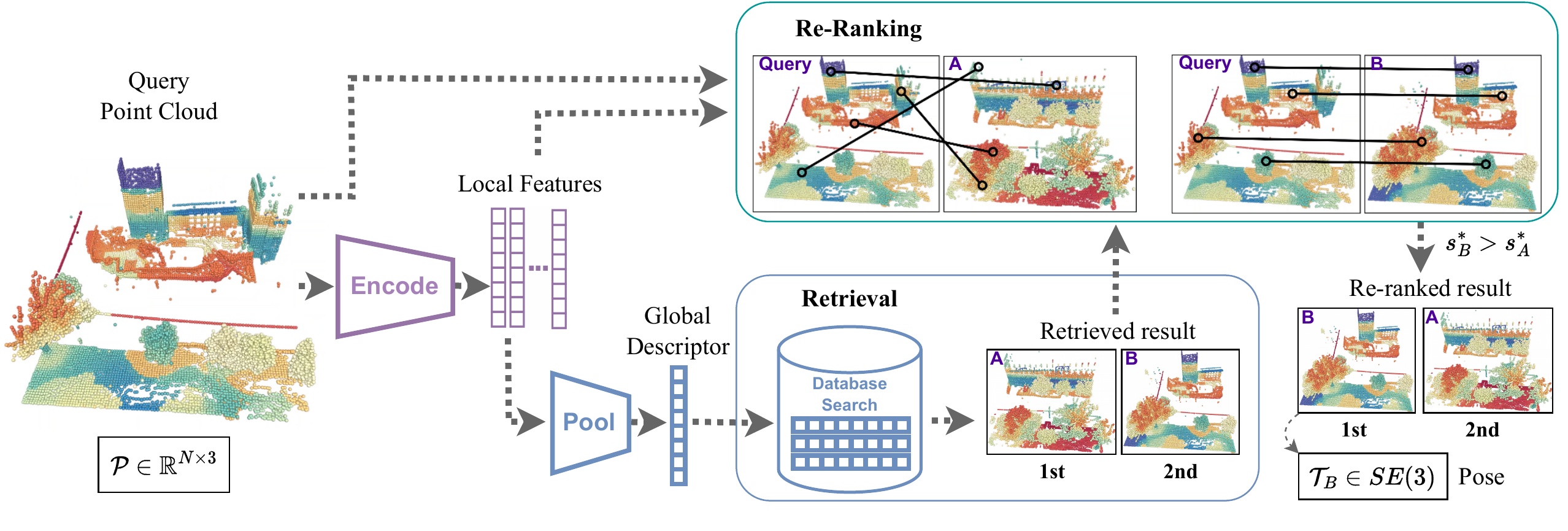}
    \caption{Overview of our SpectralGV method. Local features are extracted from the query point cloud using an encoder and then aggregated to form a global descriptor which is used in the `Retrieval' block to find the top-k place candidates (k=2 in the diagram). The local features and point cloud are used for  re-ranking the retrieved candidates in the `Re-Ranking' module which estimates a matching confidence based on the spatial consistency of point correspondences (black lines). 
    In this example, retrieved point cloud `B' is more consistent with the query and is therefore ranked higher than `A' after re-ranking.
    The 6DoF pose estimate $ \mathcal{T_B} $ is obtained by aligning the query point cloud to the top re-ranked candidate. Point clouds are coloured based on the t-SNE embeddings of local features to better visualize the point correspondences. 
    }
    \label{fig:bl_pipeline}
\end{figure*}

%% file: chapters/related_work.tex
\section{Related Work}
\label{sec:rel_work}

Point cloud retrieval methods have mainly been proposed under the two topics of place recognition and metric localization, which we discuss in \ref{sec:rw_pr} and \ref{sec:rw_met_loc} respectively. We review re-ranking methods used in other domains in \ref{sec:rw_rerank} and discuss them in relation to our proposal.

\subsection{Place Recognition}
\label{sec:rw_pr}

Large-scale place recognition is formulated as a retrieval problem where methods encode point clouds to a compact global descriptor to be used for querying a database of previously visited places. Global descriptors can be categorised as handcrafted \cite{kim2018scan, scancontextppTRO}, hybrid \cite{vidanapathirana2020locus}, and end-to-end learning \cite{uy2018pointnetvlad, liu2019lpdnet, Komorowski_2021_WACV, ma2022overlaptransformer, knights2022incloud}. While handcrafted methods such as ScanContext \cite{kim2018scan} and hybrid methods such as Locus\cite{vidanapathirana2020locus} still act as strong baselines, end-to-end learning methods have demonstrated superior performance \cite{vid2022logg3d, knights2023wildplaces}. 
End-to-end learning methods define a neural network to map point clouds to global descriptors. 
PointNetVLAD\cite{uy2018pointnetvlad} pioneered the end-to-end trainable global descriptor by combining PointNet~\cite{Qi_2017_CVPR} and NetVLAD ~\cite{arandjelovic2016netvlad}. Works such as LPD-Net \cite{liu2019lpdnet} and MinkLoc3D\cite{Komorowski_2021_WACV} have addressed the limitations of PointNetVLAD. 
Recently, LoGG3D-Net\cite{vid2022logg3d} demonstrated the benefits of using joint constraints on the local and global embeddings during the training. 
In this paper, we demonstrate how the recall of these methods can be improved by a large margin 
by our re-ranking method.

\subsection{Metric localization}
\label{sec:rw_met_loc}

DH3D \cite{du2020dh3d} pioneered correspondence-based metric localization using LiDAR data by proposing the unification of global place recognition with local 6DoF refinement. They used a simultaneous detect-and-describe formulation to extract local features, NetVLAD   to generate a global descriptor, and RANSAC\cite{fischler1981ransac}  to estimate the 6DoF pose. 
LCDNet\cite{cattaneo2022lcdnet} proposed 
the use of a  differentiable relative pose head based on Unbalanced Optimal Transport (UOT) \cite{chizat2018uot} to match local features. %
EgoNN\cite{egonn} extended the MinkLoc3D architecture for the task of metric localization with the addition of keypoint detection and saliency prediction modules. Global descriptors formed by GeM\cite{radenovic2018cnnretrieval} pooling were used for retrieval and RANSAC was used for pose estimation.

Our proposed SpectralGV can be readily integrated into all the above metric localization methods \cite{du2020dh3d, cattaneo2022lcdnet, egonn } as well as place recognition methods such as \cite{vid2022logg3d} which produce discriminative local features. 
We demonstrate how this integration results in an improvement of both recall and pose estimation for all architectures across all datasets, while having an insignificant effect on their runtime.

\subsection{Re-ranking}
\label{sec:rw_rerank}

In many retrieval problems, re-ranking methods take the initial retrieved candidates and re-order them such that correct candidates will be ranked higher. These methods can be categorised as those which use only the global descriptors and those which use both global and local embeddings. Re-ranking methods which use only global descriptors can be classified under Query Expansion (QE) \cite{radenovic2018cnnretrieval} methods that use first-order neighbour information and methods that use higher-order neighbour information such as k-reciprocal \cite{qin2011hello, zhong2017re} and region-diffusion \cite{iscen2017efficient}. 
QE methods  take the nearest neighbours of the query and generate an updated query which is used to retrieve a new set of candidates. The updated query is an aggregation of the initial top-candidates (\eg{} average-QE, alpha-QE\cite{radenovic2018cnnretrieval}). 
Higher-order methods consider the neighbours-of-neighbours of the query and are non-trivial to extend to large-scale scenarios \cite{tan2021rerankingtransformer}.
In this paper, we demonstrate how such methods which only use global information are not robust in the point cloud domain.

As opposed to general information retrieval tasks (\eg{} text-retrieval), when re-ranking in the context of localization (using either images or point clouds), the geometric cues present in the scene can be used to aid the re-ranking. For this, the local features can be used in addition to the global descriptors to verify geometric consistency between the query and the retrieved candidates. In this paper, we study how such geometric verification can be performed on point clouds by adapting methods from other fields and also introduce a more efficient method for this purpose.

Geometric verification based re-ranking has become popular for visual place recognition \cite{cao2020delg,hausler2021patchnetvlad,lee2022correlationverification}. DELG \cite{cao2020delg} used RANSAC\cite{fischler1981ransac} based geometric verification on the local features for re-ranking. 
PatchNetVLAD\cite{hausler2021patchnetvlad} compared RANSAC based re-ranking with a novel rapid-spatial-scoring mechanism for efficient geometric verification on images.  
In this paper, we study how RANSAC and newer geometric estimators \cite{zhou2016fgr, yang2020teaser} can be adapted for performing geometric verification on point clouds, and how they are inefficient when re-ranking a large number of potential candidates. We propose a geometric verification method with sublinear time complexity to address this drawback.

%% file: chapters/method.tex
\section{Proposed Method}
\label{sec:method}

This section describes the re-ranking based metric localization pipeline depicted in ~\figref{fig:bl_pipeline} and presents our SpectralGV method for efficient geometric verification.

\subsection{Problem Formulation}
\label{section:formulation}

Our approach to metric localization is split into 3 sub-tasks: retrieval, re-ranking and 6DoF pose estimation. 

\textbf{Retrieval: } The sub-task of point cloud retrieval is illustrated in the `Retrieval' module of ~\figref{fig:bl_pipeline} and formulated as follows. Given a point cloud $ \mathcal{P} \in \mathbb{R}^{N \times 3} $ representing varying number of $N$ 3D points,
a mapping function $\Phi$ such that 
\begin{equation}
\label{eqn:mapping_function}
    \Phi : \mathcal{P} \rightarrow (\mathcal{F}, g)
\end{equation}
is learnt to represent the point cloud $ \mathcal{P} $ as a set of local features $ \mathcal{F} \in \mathbb{R}^{N \times d'}$,  and also as a fixed-size vector global descriptor $ g \in \mathbb{R}^{d} $.
We refer the reader to \cite{vid2022logg3d} for a detailed description of how $ \Phi $ can be learnt to produce discriminative local and global embeddings.
During retrieval, the global descriptor $ g $ of the query point cloud  $ \mathcal{P} $ is matched with a database of global descriptors of previously visited places to obtain
the list of top-k retrieval candidates,
$
    \mathcal{L}_{R} = \{ \mathcal{P}_{R_{1}} , ... , \mathcal{P}_{R_{k}} \}, 
$
where $ \mathcal{L}_{R} $ is ordered  based on  similarity of retrieved global descriptors with $ g $. If not employing re-ranking, $ \mathcal{P}_{R_{1}} $ is taken as the solution for place recognition. 

\textbf{Re-Ranking: } In the second sub-task depicted in the `Re-Ranking' module of ~\figref{fig:bl_pipeline}, geometric verification based re-ranking techniques aim to calculate a  geometric `fitness score'  ($ s $) between the query and each retrieved candidate, 
  where the score measures the spatial consistency between the query and each candidate using point correspondences estimated by matching the local features $ \mathcal{F} $.

The initial top-k retrieval candidates are then re-ranked by sorting based on descending fitness score to obtain, 
\begin{equation}\label{eqn:reranking}
    \mathcal{L}_{RR} = \{ \mathcal{P}_{RR_{1}} , ... , \mathcal{P}_{RR_{k}} \},
\end{equation}
where the list $ \mathcal{L}_{RR} $ is a permutation/re-ordering of the list $ \mathcal{L}_{R} $. A successful re-ranking method will ensure that,
\begin{equation}\label{eqn:problem_1}
  \mathcal{D}(\mathtt{x},\mathtt{x_{RR_{i}}}) \leq  \mathcal{D}(\mathtt{x},\mathtt{x_{R_{i}}}) \quad \forall i,
\end{equation}
for all queries where $\mathtt{x}, \mathtt{x_{R_{i}}}, \mathtt{x_{RR_{i}}}$ are the ground-truth geo-locations of $ \mathcal{P}, \mathcal{P}_{R_{i}}, \mathcal{P}_{RR_{i}} $ respectively, and  $ \mathcal{D} $ represents Euclidean distance. 
We note that re-ranking methods which are not robust will occasionally violate this inequality (due to failure cases such as noisy inputs, degenerate scenes). In some instances, they will reduce retrieval performance after re-ranking. We empirically demonstrate how all geometric verification methods increase performance after re-ranking and how our SpectralGV shows the biggest improvement while being the only geometric verification method with sub-linear time complexity, hence enabling its real-time use.

\textbf{Pose Estimation: } During the third sub-task, the pose estimate is obtained by registering the query $ \mathcal{P} $ with the top-1 re-ranked candidate $ \mathcal{P}_{RR_{1}} $ to obtain their relative transformation: $ \mathcal{T}_{RR_{1}} \in SE(3) $.
 This transformation can be estimated using any point cloud registration technique. Typically, for LiDAR-based metric-localization, since local features and global descriptors are extracted concurrently, correspondence-based registration techniques using robust estimators such as RANSAC are preferred over alternatives.

 \subsection{Spectral Geometric Verification}

  This paper focuses on the sub-task of re-ranking, and the proposed method can be readily incorporated with existing architectures which support the formulation presented in Eq.\ref{eqn:mapping_function} - global descriptor based retrieval and local feature correspondence based registration. Baseline 
  geometric verification methods
  first register the query with each of the retrieved point clouds, calculate the point inlier-ratio after registration, and then sort the retrievals based on the Registered-Inlier-Ratio (RIR). This type of re-ranking is inefficient as it requires registering the query with each retrieval.
  In this paper, we introduce a spectral fitness score as an alternative to the RIR for geometric verification based re-ranking.

Inspired by the Spectral Matching method introduced in \cite{spectralmatch}, 
we define an un-directed graph $ \mathcal{G} =  \mathcal{(V,E)} $ between the query $ \mathcal{P} $ and each retrieved candidate $ \mathcal{P}_{R} \in \, \mathcal{L}_{R} $.
Vertices $ \mathcal{V} = \{1, .., n\} $ represent the $ n $  elements in the set of local point correspondences $ \mathcal{C} $  between the two point clouds obtained by nearest-neighbour search on their local features. Here $ {c_i} \in \, \mathcal{C} $ is denoted as ${c_i} = (x_i, y_{i}) $, where $ x_i \in \, \mathcal{P}  , y_{i} \in \, \mathcal{P}_{R} $ are the corresponding points from the two point clouds. 
The edges $ \mathcal{E} \subseteq \mathcal{V} \times \mathcal{V} $ represent the spatial compatibility score of these correspondences (defined later). The motivation here is that this graph will form clusters of nodes which have high spatial consistency with each other, and the largest cluster of this graph is statistically likely to be the inlier correspondences, as it is highly unlikely that outlier correspondences will be spatially consistent with the inliers or with each other.

\input{chapters/tables/method_image.tex}

The adjacency matrix of this graph represented as
$ M \in \mathbb{R}^{n \times n} $ is a symmetric matrix of all positive elements where each element $ m_{i,j} $ represents a spatial compatibility score between the correspondences $c_i, c_j$. Specifically, $m_{i,j}$ measures how well the pairwise geometry of two points in the query point cloud are preserved after matching them with two points in the retrieved point cloud. This is defined as,
\begin{equation}
    m_{i,j} = [1 - \frac{d_{ij}^2}{d_{thr}^2}]_+,  d_{ij} = \big | \left\Vert {x_i-x_j}\right\Vert - \left\Vert {y_i-y_j}\right\Vert \big |,
\label{eq:spatial_consistency}    
\end{equation}

where $[\cdot]_+ = \max (\cdot, 0)$ ensures a non-negative value for $m_{ij}$, 
and $d_{thr}$ controls the sensitivity to the length difference. As depicted in Fig. \ref{fig:method_image}, correspondence pairs with length difference (of dashed lines) larger than $d_{thr}$ (\emph{e.g.} ${c_1, c_2}$) are considered incompatible and get a zero value in $M$, while pairs which preserve their length relationships (\emph{e.g.} ${c_1, c_3}$ obtain high positive scores. 

The task of finding the inlier correspondences is formulated as finding the cluster $ \hat{C} \subseteq \, \mathcal{C}  $ of correspondences that maximize the inter-cluster spatial compatibility score,
\begin{equation}
\label{eqn:cluster_score}
    s = \sum_{c_{i},c_{j} \in \hat{C}} m_{i,j},
\end{equation}
subject to the constraint that no pairs of incompatible correspondences are present in $ \hat{C} $. By representing the cluster $ \hat{C} $ using a binary indicator vector $ v $ such that $ v(c_i) = 1 $ if $ c_i \in \hat{C} $ and zero otherwise, we can re-write the inter-cluster score in Eq.\ref{eqn:cluster_score} as $ s = v^{\intercal}Mv$. Given this formulation the indicator vector that obtains the maximum inter-cluster score can be computed as, 
\begin{equation}
    v^{*} = \argmax_{v} \, (v^{\intercal}Mv).
\label{eq:v_max} 
\end{equation}
As demonstrated in \cite{spectralmatch}, the compatibility constraint and the integer constraint on $v$ can be relaxed and it's norm can be fixed to 1 since only the relative values of it's elements are important. Then,
Eq. \ref{eq:v_max} can be represented as, 
\begin{equation}
    v^{*} = \argmax_{v} \left(  \frac{v^{\intercal}Mv}{v^{\intercal}v} \right) 
\label{eq:final} 
\end{equation}
where $v^{*}$ can be approximated using the principal eigenvector of $M$ according to the Rayleigh quotient. We note that obtaining an exact solution for $v^{*}$ accounts to finding the inlier correspondences given initial noisy correspondences and this problem does not have a closed form solution. While the methods of Spectral Matching \cite{spectralmatch} and its recent variants \cite{bai2021pointdsc} have used $v^{*}$ to solve the registration problem, in this paper, we explore a different direction. We interpret the correspondence compatibility graph as a graphical representation of the geometric compatibility of the two point clouds, and thus a scalar-value summary of this compatibility can be obtained via the optimal inter-cluster score. With this interpretation, we demonstrate (for the first time) that a fitness score for geometric verification based re-ranking can be obtained without registration of the two point clouds. \\

The maximum inter-cluster score (\ie{} the sum of spatial compatibility scores $ m_{i,j} $ of the approximated inlier correspondences) is obtained using:
\begin{equation}
 s^{*} = v^{*\intercal}Mv^{*}, 
\end{equation}
 and can be interpreted as a scalar value summarizing the geometric consistency between two point clouds and therefore can be used as a fitness score for geometric verification based re-ranking. Additionally, as demonstrated in \cite{spectralmatch}, the approximation of $v^{*}$  in this manner is statistically robust against outlier correspondences.
Hence, we postulate that this robustness of $v^{*}$ naturally translates to the score $s^{*}$ which results in a robust geometric verification method enabling the preservation of the inequality in \ref{eqn:problem_1}. This robustness for the task of re-ranking substantiated by our extensive experimental results.

Using this interpretation of $s^{*}$, $ \mathcal{L}_{RR} $ can be obtained by sorting $ \mathcal{L}_{R} $ based on descending $s^{*}$ calculated between the query $ \mathcal{P} $ and each  $ \mathcal{P}_{R} \in \, \mathcal{L}_{R} $
, using Eq.\ref{eqn:reranking}. This process is illustrated in ~\figref{fig:bl_pipeline} for the case of 2 candidates where the initial top-1 candidate `A' is replaced by the 2nd candidate `B' after re-ranking because $ s^{*}_B > s^{*}_A $. Additionally, we provide an implementation that parallelizes this computation across all retrieved candidates, enabling robust large-scale point cloud re-ranking in realtime.

%% file: chapters/tables/method_image.tex
\begin{figure}[t]
\centering
\includegraphics[width=0.49\textwidth]{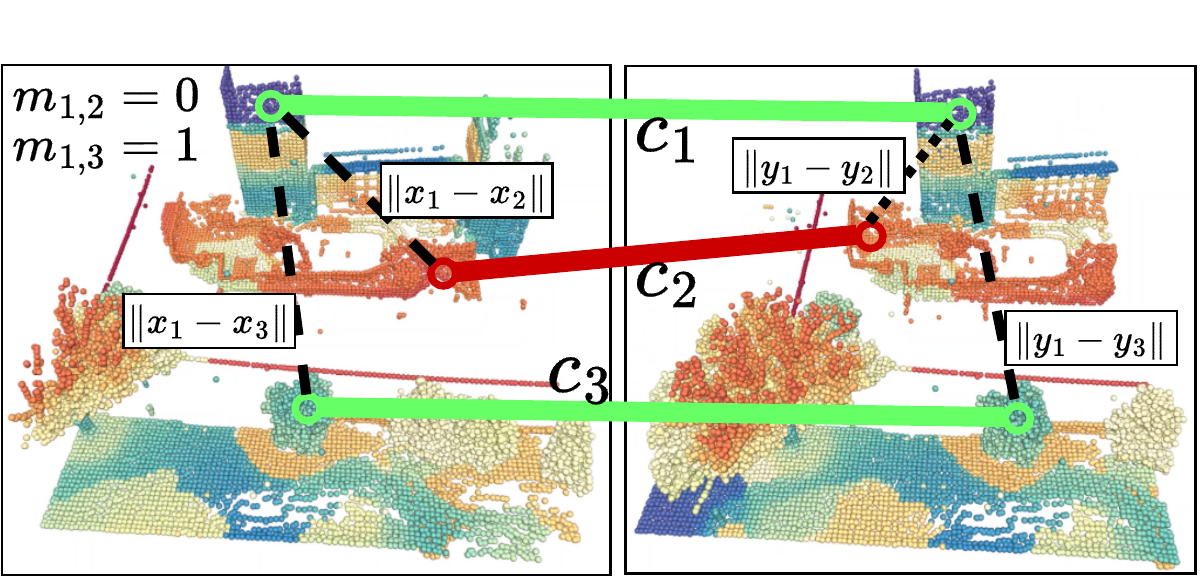}
\caption{Visualization of values in the adjacency matrix $ M $. The colored lines represent inter-pointcloud correspondences. Out of the 3 depicted correspondences, the green ones ($c_1, c_3 $) are inliers and the red ($c_2 $) are outliers. The compatibility of these  correspondences are represented by the $m_{i,j}$ values in the top left of this figure.  Inlier correspondences will preserve their spatial consistency (low $d_{1,3}$) and form strong edges in $ \mathcal{G} $ (high $m_{1,3}$), while outliers will not form strong edges with the inliers (low $m_{1,2}$).}
\label{fig:method_image}
\vspace{-5mm}
\end{figure}

%% file: chapters/experiments.tex
\section{Experiments}
\label{sec:experiments}
We conduct extensive experiments to demonstrate that SpectralGV (1) outperforms other re-ranking methods, and (2) improves performance for all architectures and datasets. %
\subsection{Experiment setup}

We integrate SpectralGV with the architectures EgoNN
\cite{egonn}, LCDNet
\cite{cattaneo2022lcdnet} and LoGG3D-Net \cite{vid2022logg3d}. 
For EgoNN we use the pre-trained model released by the authors which was trained in the same train set as used in this paper. LoGG3D-Net and LCDNet are trained on a cluster of P100-16GB GPUs. LCDNet was trained for 150 epochs with batch size of 4 using identical hyper-parameters from \cite{cattaneo2022lcdnet}. 
 We use the power-iteration method to efficiently approximate the leading eigenvector $v^{*}$ in Eq. \ref{eq:final}.

\input{chapters/tables/table_data_splits}

We evaluate on five public LiDAR datasets, using an identical training set as in \cite{egonn}: MulRan~\cite{mulran20} and Apollo-SouthBay\cite{apollo_dataset} datasets. We define 2 evaluation settings (`easy' and `hard') where the easy setting consists of the test sequences proposed in \cite{egonn},
and for the hard setting we introduce additional sequences (unseen during training) in which current methods struggle to generalize: MulRan DCC~\cite{mulran20}, KITTI-360~\cite{geiger2013kittidataset} and the ALITA~\cite{yin2022alita} datasets. In  MulRan DCC, sequences 01 and 02 are used as database and query sets. In KITTI-360, sequence 09 is used with the first 300 seconds of the sequence as a database and the rest as queries. In ALITA, we evaluate on the data released at the ICRA 2022 UGV Challenge  and use the validation sequence 5.  Details on the number of scans and sampling step lengths of each dataset are given in Table.~\ref{tab:data_splits}.

\subsection{Evaluation Criteria}

For place recognition, we compute the similarity between the global descriptors of each query point cloud with global descriptors from the database set and get the top-k retrieval candidates. 
We report Recall@k (Rk) (for k = $1$ and $5$) metrics for $5$ and $20$\,m revisit thresholds.
We also use the Mean Reciprocal Rank (MRR) as it serves as a scalar summary of the Recall@k curve defined as  $MRR=\frac{1}{|Q|}\sum_{q \in Q} \frac{1}{r_{q}}$ where $ r_q $ is the rank of the first true-positive retrieval for a given query $ q $. We report both metrics as percentages.

We first retrieve the top candidates for each query and calculate the baseline R1, R5 and MRR results (presented in the 1st row of table ~\ref{tab:rerank_egonn} and the top-halves of tables ~\ref{tab:place_recog_easy} and ~\ref{tab:place_recog_hard}).
Then we re-rank the $ n_{topk} $ retrieval candidates and recalculate R1, R5 and MRR.
We use $ n_{topk} = 20 $ for the evaluations in the bottom halves of Table.~\ref{tab:place_recog_easy} and ~\ref{tab:place_recog_hard}.

For metric localization,  the 6DoF pose estimate of each method is compared to the ground truth.
The success rate (S) is the percentage of pose estimates within 2\,m and $5^{\circ}$ threshold from the ground truth.
The mean Relative Rotation Error (RRE) and  mean Relative Translation Error (RTE) metrics are as defined in \cite{bai2021pointdsc} and are calculated for all queries (not only the successful queries as done in \cite{egonn}).

%% file: chapters/tables/table_data_splits.tex
\begin{table}[t]

\begin{center}
\begin{tabular}{@{\enspace}l@{\enspace}|@{\enspace}r@{\enspace}|@{\enspace}c@{\enspace}|@{\enspace}c@{\enspace}}
\begin{tabular}{@{}c@{}} Train/Test Split \end{tabular}
& \begin{tabular}{@{}c@{}}Length \end{tabular}
& \begin{tabular}{@{}c@{}}Step \end{tabular}
& \begin{tabular}{@{}c@{}}Number of scans\end{tabular}
\\[2pt]
\hline
\textit{Train split:}  &  &  \\
MulRan \cite{mulran20}: Sejong (0.1/02) & 19 km & 0.2 m & 35,871 \\
SouthBay \cite{apollo_dataset}: excl. Sunnyvale  &  37 km & 1.0 m & 72,706 \\
\hline
\textit{Test split (Easy):}  &  & & (database\,/\,query) \\
MulRan\cite{mulran20}: Sejong (01/02)  &  4 km & 0.2 m & 3,764\,/\,3,453 \\
SouthBay\cite{apollo_dataset}: Sunnyvale  &  38 km & 1.0 m & 48,959\,/\,16,999  \\
KITTI\cite{geiger2013kittidataset}: Sequence 00 & 4 km & 0.1 m & 1,620\,/\,621  \\
\textit{Test split (Hard):}  &  &  \\
MulRan\cite{mulran20}: DCC (01/02)  & 4.9 km & 10.0 m & 469\,/\,307  \\
ALITA\cite{yin2022alita} (val:5)  & 1.4 km & 3-20 m & 107\,/\,84  \\
KITTI-360\cite{liao2022kitti360}: Sequence 09  & 10.5 km & 3.0 m & 766\,/\,398  \\
\end{tabular}
\caption{Details of training and evaluation sets.}
\label{tab:data_splits}
\end{center}
\vspace{-6mm}
\end{table}

%% file: chapters/results.tex
\subsection{Results}
\label{sec:results}

\subsubsection{Re-ranking evaluation}

\input{chapters/tables/ntopk_plot.tex}
\input{chapters/tables/table_rerank_egonn}
\input{chapters/tables/qualitative_rerank.tex}

We explore different methods for re-ranking of point cloud retrievals. We evaluate on the `hard' evaluation sequences and report Recall@1, MRR metrics and the time taken `t' (in ms) for re-ranking in Table. \ref{tab:rerank_egonn} using the EgoNN architecture. We present results under two settings: using 2 and 20 $ n_{topk} $ candidates for re-ranking.

We group the re-ranking methods into 3 categories. The first category (group (a) in Tab. \ref{tab:rerank_egonn}) only uses the global descriptor. We evaluate two Query Expansion (QE) methods: Average-QE and Alpha-QE \cite{radenovic2018cnnretrieval}. We observe that Average-QE does not produce an improvement in any setting. 
This is because majority of the candidate descriptors in the initial retrieval are incorrect matches, hence averaging these in to the new query simply produces a noisy query. Alpha-QE shows some improvements in Recall@1 when $n_{topk} = 2$, but has degrading performance with increasing $n_{topk}$. 
While global-only methods are efficient in point cloud re-ranking, they lack robustness and perform worse than the baseline.  

The second category consists of geometric verification (GV) methods that re-rank based on the registered-inlier-ratio (RIR). We test 3 robust geometric estimators to perform the registration: RANSAC\cite{fischler1981ransac}, FGR\cite{zhou2016fgr} and TEASER++\cite{yang2020teaser}. For RANSAC and FGR we use the efficient implementations provided in \cite{zhou2018open3d}. %
We observe that all RIR methods improve performance in all settings. RANSAC obtains the best results in the RIR category (group (b) in Tab. \ref{tab:rerank_egonn}). The final category of re-ranking methods (group (c) in Tab. \ref{tab:rerank_egonn}) consists of geometric verification methods that do not perform registration 
but directly compute a spatial consistency (SC) score. SpectralGV falls into this category and we observe that it outperforms all other re-ranking methods in Tab. \ref{tab:rerank_egonn}. 

All geometric verification methods show increasing recall with increasing  $n_{topk}$. RIR methods significantly increase  runtime with increasing $n_{topk}$ (increasing by an order of magnitude in all cases when increasing $n_{topk}$ from 2 to 20) but SpectralGV maintains roughly constant computation time. In the $n_{topk}$=20 setting, our method obtains the best recall overall while also being more efficient than other GV methods as shown in the final row of Tab. \ref{tab:rerank_egonn} operating at 3-5\,ms while other GV methods take 50-300\,ms. 
The effect of $n_{topk}$ on re-ranking performance and on efficiency for the best performing method in each category is presented in Fig. \ref{fig:ntopk_plot} using EgoNN on KITTI-360 09 under the 5\,m revisit criteria. Only GV methods show consistently increasing Recall@1. While RIR methods have linearly increasing runtime and become impractical at large $n_{topk}$ values, SpectralGV maintains almost constant runtime while also outperforming RIR methods in recall.

We conduct a qualitative evaluation of re-ranking results on the 5 datasets. Fig. \ref{fig:quali_rerank} shows point clouds corresponding to the query, initial top-1 retrieval and re-ranked top-1 retrievals in cases where an incorrect initial candidate is corrected after re-ranking. In all instances, the incorrect initial candidate point cloud is highly structurally similar to the query with only minor inconsistencies visible upon close inspection. Our re-ranking method is able to handle such complex cases and accurately identify the correct candidate. %

Re-ranking enables better use of test-time knowledge. Current retrieval methods still struggle to discriminate between complex scenes which have largely similar structural elements as they only rely on global embeddings. These ambiguities can be addressed at test time using re-ranking of the top retrieved candidates as the correct candidate is more likely to be within the top-k retrievals than top-1.

\input{chapters/tables/table_placog}
\input{chapters/tables/table_placog_hard}

\subsubsection{Place recognition evaluation}

We compare place recognition performance using the handcrafted method ScanContext~\cite{kim2018scan}, hybird method Locus~\cite{vidanapathirana2020locus} and end-to-end learning methods MinkLoc3D~\cite{Komorowski_2021_WACV}, DiSCO~\cite{xu2021disco}, DH3D~\cite{du2020dh3d}, EgoNN~\cite{egonn}, LCDNet~\cite{cattaneo2022lcdnet} and LoGG3D-Net~\cite{vid2022logg3d}.
Table~\ref{tab:place_recog_easy} shows place recognition results on the `easy' evaluation sets while Table.~\ref{tab:place_recog_hard} shows the corresponding results in the `hard' evaluation sequences.
On the `easy' evaluation set, we see that LoGG3D-Net outperforms the previous state-of-the-art EgoNN on average performance across all evaluation metrics. For the methods which also output local features (EgoNN, LCDNet and LoGG3D-Net), we also evaluate performance after re-ranking. 
The results on the `easy' set are already saturated prior to re-ranking but we still see improvements (or on-par results) for all methods after re-ranking using SGV. 
LoGG3D-Net + SGV obtains the best performance overall on the `easy' datasets.

On the `hard' evaluation setting all methods do worse as not only are these sequences unseen during training, but they also contain difficult evaluation scenarios such as reverse/orthogonal revisits and lane offsets during revisits. 
Therefore, the performance gain after re-ranking is much more pronounced on the `hard' evaluation set. We observe that LoGG3D-Net which fails to generalize to the ALITA dataset shows an improvement of 32.5 MRR after re-ranking which is more than double the initial MRR while also almost quadrupling R1 after re-ranking. 
The results after re-ranking using our SGV are better for all methods in all datasets. This demonstrates that our method improves retrieval performance on all datasets while being architecture agnostic.

\subsubsection{Metric localization evaluation}

\input{chapters/tables/table_metloc_rerank.tex}

We evaluate LCDNet\cite{cattaneo2022lcdnet}, EgoNN\cite{egonn} and LoGG3D-Net\cite{vid2022logg3d}. All methods use RANSAC for pose estimation. Although LCDNet uses unbalanced optimal transport for transformation estimation during training, RANSAC is used during inference \cite{cattaneo2022lcdnet}.

Table~\ref{tab:met_loc_new} shows 6DoF pose estimation results. 
We observe that prior to re-ranking, all 3 methods have large RTE (close to or greater than 10\,m), limiting their reliability for re-localization applications. After re-ranking using SpectralGV (with $ n_{topk} = 20 $), all 3 methods show significant performance improvements. For EgoNN, the RTE and RRE have reduced by 89.9\% and 89.2\% respectively. The RTE of LCDNet and LoGG3D-Net have reduced by 92.4\% and 69.4\% respectively. All methods now have RTE $< 5\,m$ and RRE $< 3^\circ$, making them more reliable for re-localization.

SpectralGV can be used to refine the retrieval candidates such that the top-1 retrieval used for registration is closer to the the query. For example, in Fig. \ref{fig:qual_rr_kitti00}, we see that SpectralGV has replaced the initial candidate which is 21\,m away from the query with a candidate which is only 3\,m away from the query. This will result in an easier registration task and hence enable more accurate metric localization results.

%% file: chapters/tables/ntopk_plot.tex
\begin{figure}[t]
\centering
\includegraphics[width=0.99\linewidth]{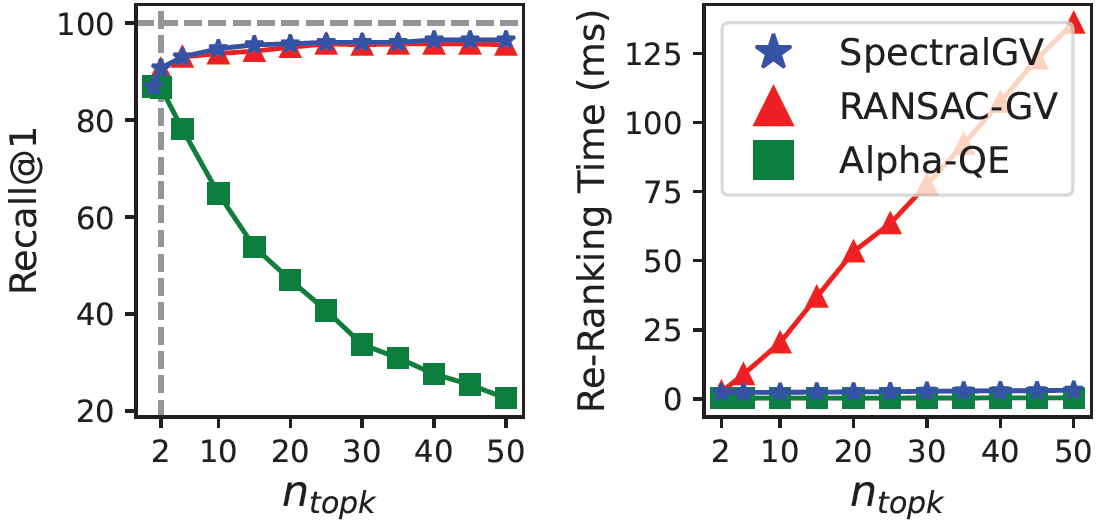}
\caption{Analysis of re-ranking methods with varying number of candidates used for re-ranking: $ n_{topk} $.
SpectralGV shows consistently  increasing Recall@1 (left plot) while maintaining sub-linear time complexity (right plot).}
\label{fig:ntopk_plot}
\vspace{-5mm}
\end{figure}

%% file: chapters/tables/table_rerank_egonn.tex
\begin{table*}[t]

\begin{center}
\begin{tabular}{l@{\quad}|c@{\quad}|c@{\quad}c@{\quad}c@{\quad}c@{\quad}c@{\quad}|c@{\quad}c@{\quad}c@{\quad}c@{\quad}c@{\quad}|c@{\quad}c@{\quad}c@{\quad}c@{\quad}c}
& &  \multicolumn{5}{c}{MulRan DCC}  & \multicolumn{5}{c}{ALITA}  & \multicolumn{5}{c}{KITTI-360} \\
& &  \multicolumn{2}{c}{5m} &  \multicolumn{2}{c}{20m} &  \multicolumn{1}{c}{}  &  \multicolumn{2}{c}{5m} &  \multicolumn{2}{c}{20m} &  \multicolumn{1}{c}{}  &  \multicolumn{2}{c}{5m} &  \multicolumn{2}{c}{20m} &  \multicolumn{1}{c}{} \\
& \begin{tabular}{@{}c@{}}$n_{topk}$\end{tabular}
& \begin{tabular}{@{}c@{}}R1\end{tabular}
& \begin{tabular}{@{}c@{}}MRR\end{tabular}
& \begin{tabular}{@{}c@{}}R1\end{tabular}
& \begin{tabular}{@{}c@{}}MRR\end{tabular}
& \begin{tabular}{@{}c@{}}t\end{tabular}
& \begin{tabular}{@{}c@{}}R1\end{tabular}
& \begin{tabular}{@{}c@{}}MRR\end{tabular}
& \begin{tabular}{@{}c@{}}R1\end{tabular}
& \begin{tabular}{@{}c@{}}MRR\end{tabular}
& \begin{tabular}{@{}c@{}}t\end{tabular}
& \begin{tabular}{@{}c@{}}R1\end{tabular}
& \begin{tabular}{@{}c@{}}MRR\end{tabular}
& \begin{tabular}{@{}c@{}}R1\end{tabular}
& \begin{tabular}{@{}c@{}}MRR\end{tabular}
& \begin{tabular}{@{}c@{}}t\end{tabular}

\\[2pt]
\hline
Baseline  & - & 67.1 & 77.2 & 90.2 & 92.3 &  - & 79.8 & 87.3 & 91.7 &  94.9 & - & 86.9 & 90.1 &  88.2 & 91.4 & -  \\
\hline
\textit{(a) Global only:} &  &  &  &  &  &   &  &  &  &   &  &  &  &   &  & \\
Average-QE  & 2 & 67.4 & 76.7 & 90.2 & 91.8 &  0.2 & 79.8 & 86.5 & 91.7 &  94.3 & 0.1 & 86.9 & 89.3 &  88.2 & 90.8 & 0.2  \\
 & 20 & 35.2 & 47.6 & 54.7 & 65.0 &  0.2 & 13.1 & 27.4 & 26.2 &  44.0 & 0.1 & 46.2 & 59.4 &  49.7 & 63.8 & 0.2  \\
Alpha-QE \cite{radenovic2018cnnretrieval} & 2 & 67.4 & 76.7 & 90.6 & 92.0 &  0.2 & 81.0 & 87.1 & 92.9 &  94.9 & 0.1 & 86.7 & 89.2 &  87.9 & 90.6 & 0.2 \\
 & 20 & 34.5 & 47.4 & 54.7 & 65.0 &  0.3 & 13.1 & 28.4 & 26.2 &  45.0 & 0.2 & 47.0 & 59.9 &  50.8 & 64.4 & 0.3  \\
\hline
\textit{(b) GV: RIR}  &  &  &  &  &  &   &  &  &  &   &  &  &  &   &  & \\
RANSAC \cite{fischler1981ransac} & 2 & 71.0& 79.0 & 92.8 &  93.6 & 2.8 & 90.5 & 92.7 &  95.2 & 96.7 & 4.1 & 90.5 & 91.9 & 92.5 & 93.5  & 2.8 \\
 & 20 & \underline{72.0} & \underline{80.9} & \textbf{95.8} & \textbf{96.0} &  63.6 & \textbf{100} & \textbf{100} & \textbf{100} &  \textbf{100} & 60.4 & \underline{95.0} & \underline{96.4} &  \underline{98.5} & \underline{98.6} & 56.3  \\
FGR \cite{zhou2016fgr} & 2 & 70.4 & 78.7 & 91.2 & 92.8 &  79.7 & 89.3 & 92.1 & 95.2 &  96.7 & 54.8 & 90.2 & 91.7 &  92.2 & 93.4 & 74.4  \\
 & 20 & 70.7 & 78.9 & 92.2 & 93.2 &  175.7 & 97.6 & 98.8 & \textbf{100} &  \textbf{100} & 126.7 & 93.7 & 95.8 &  98.0 & 98.3 & 171.3  \\
TEASER++ \cite{yang2020teaser} & 2 & 70.0 & 78.5 & 92.5 & 93.4 &  26.3 & 90.5 & 92.7 & 95.2 &  96.7 & 30.1 & 89.9 & 91.6 &  92.2 & 93.4 & 35.1  \\
 & 20 & 71.3 & 80.6 & 93.8 & 94.8 &  261.3 & 98.8 & 99.4 & \textbf{100} &  \textbf{100} & 272.8 & 94.0 & 95.8 &  96.7 & 97.6 & 320.5  \\
\hline
\textit{(c) GV: SC}  &  &  &  &  &  &   &  &  &  &   &  &  &  &   &  & \\
\textbf{SpectralGV (Ours)} & 2 & 71.7 & 79.3 & 92.8 & 93.6 &  2.7 & 90.5 & 92.7 & 95.2 &  96.7 & 5.3 & 90.7 & 92.0  &  92.5 & 93.5 & 2.4  \\
 & 20 & \textbf{73.6} & \textbf{82.2} & \underline{95.1} & \underline{95.7} &  3.0 & \textbf{100} & \textbf{100} & \textbf{100} &  \textbf{100} & 5.7 & \textbf{95.7} & \textbf{96.9} &  \textbf{	98.7} & \textbf{98.8} & 2.8  \\[2pt]
\end{tabular}
\caption{
Comparison of re-ranking methods. The first row shows the baseline retrieval results without re-ranking.
Methods in subsection (a) perform re-ranking using only the global descriptors. Methods in subsections (b) and (c) perform Geometric Verification (GV) based re-ranking where (b) computes the Registered-Inlier-Ratio (RIR) and (c) directly computes a Spatial Consistency (SC) score.
 $n_{topk}$: number of top-k retrieval candidates used during re-ranking.  %
}
\label{tab:rerank_egonn}
\end{center}
\end{table*}

%% file: chapters/tables/qualitative_rerank.tex
\begin{figure*}[t]
     \centering
     \begin{subfigure}[b]{0.1949\textwidth}
         \centering
         \includegraphics[width=\textwidth]{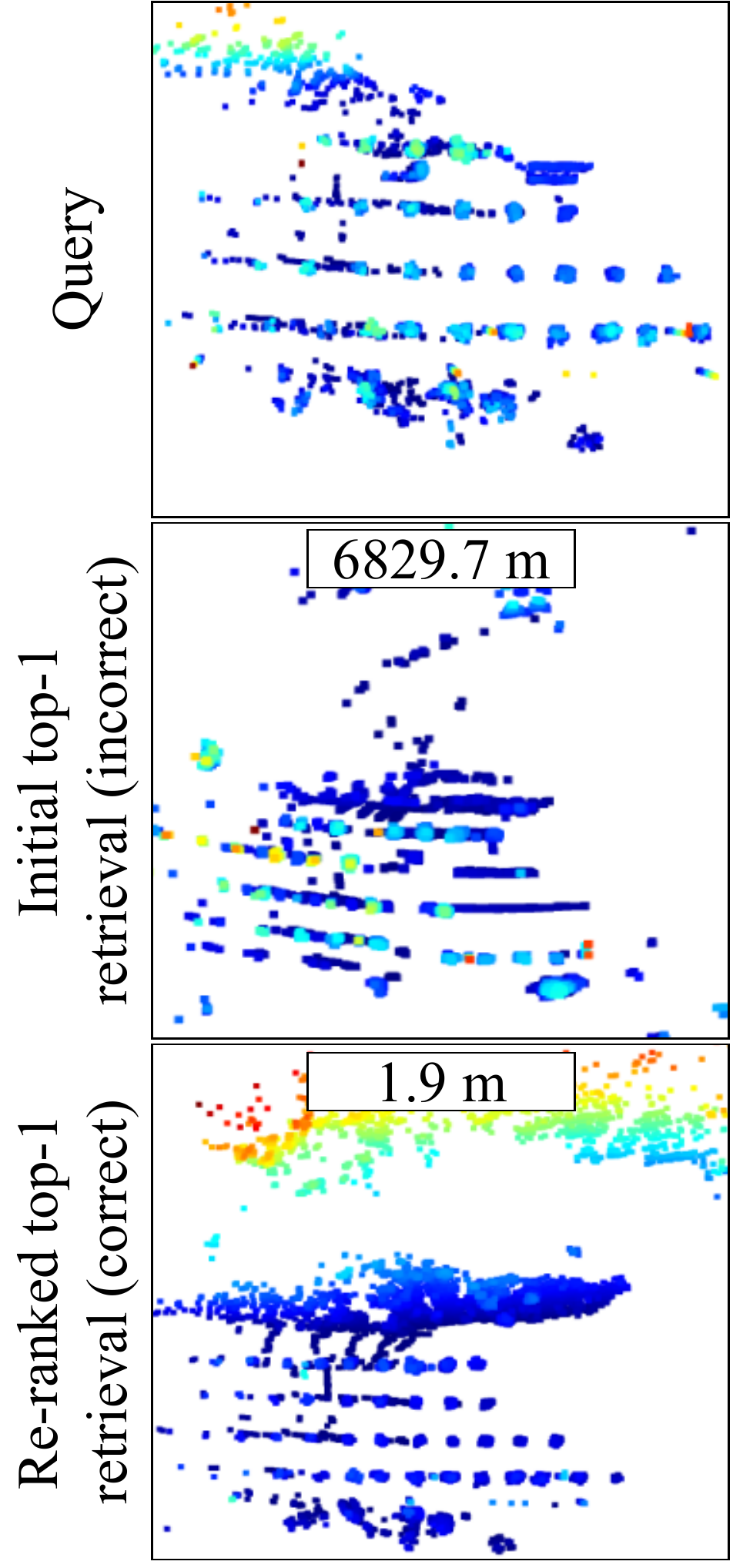}
         \caption{MulRan Sejong}
         \label{fig:qual_rr_sejong}
     \end{subfigure}
     \hfill
     \begin{subfigure}[b]{0.1549\textwidth}
         \centering
         \includegraphics[width=\textwidth]{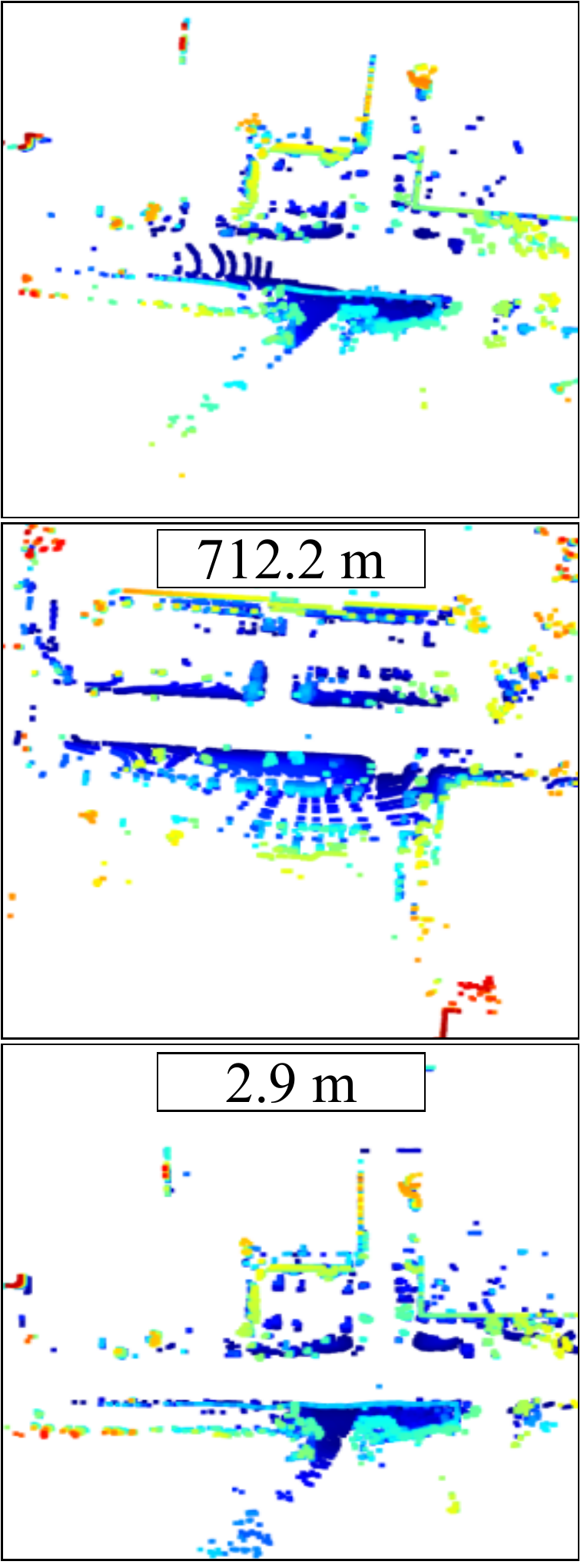}
         \caption{Apollo-Southbay}
         \label{fig:qual_rr_apollo}
     \end{subfigure}
     \hfill
     \begin{subfigure}[b]{0.1549\textwidth}
         \centering
         \includegraphics[width=\textwidth]{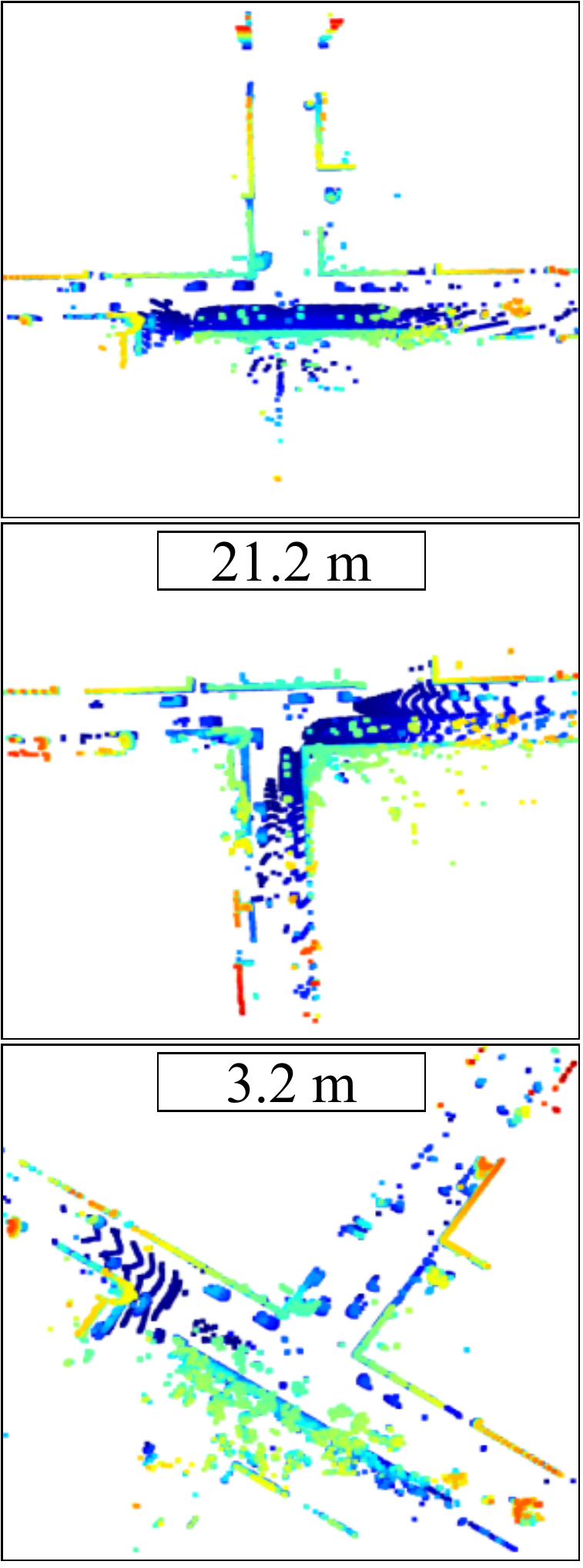}
         \caption{KITTI 00}
         \label{fig:qual_rr_kitti00}
     \end{subfigure}
     \hfill
     \begin{subfigure}[b]{0.1549\textwidth}
         \centering
         \includegraphics[width=\textwidth]{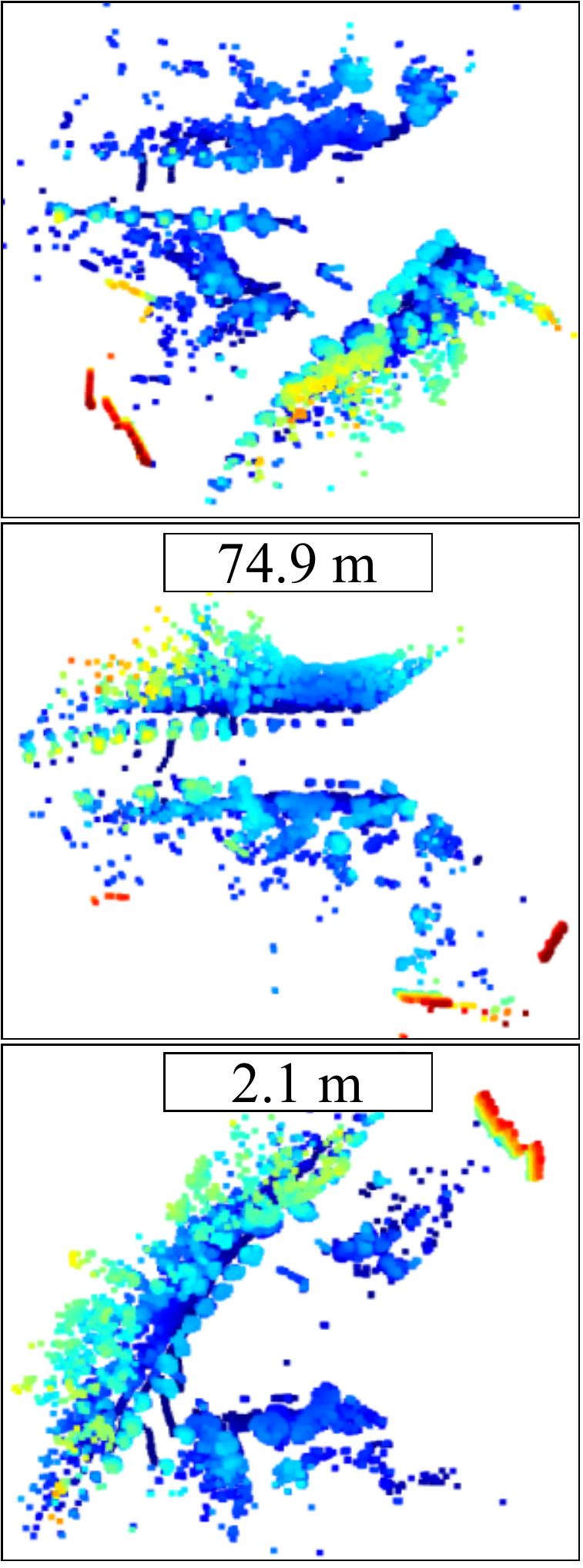}
         \caption{MulRan DCC}
         \label{fig:qual_rr_dcc}
     \end{subfigure}
     \hfill
     \begin{subfigure}[b]{0.1549\textwidth}
         \centering
         \includegraphics[width=\textwidth]{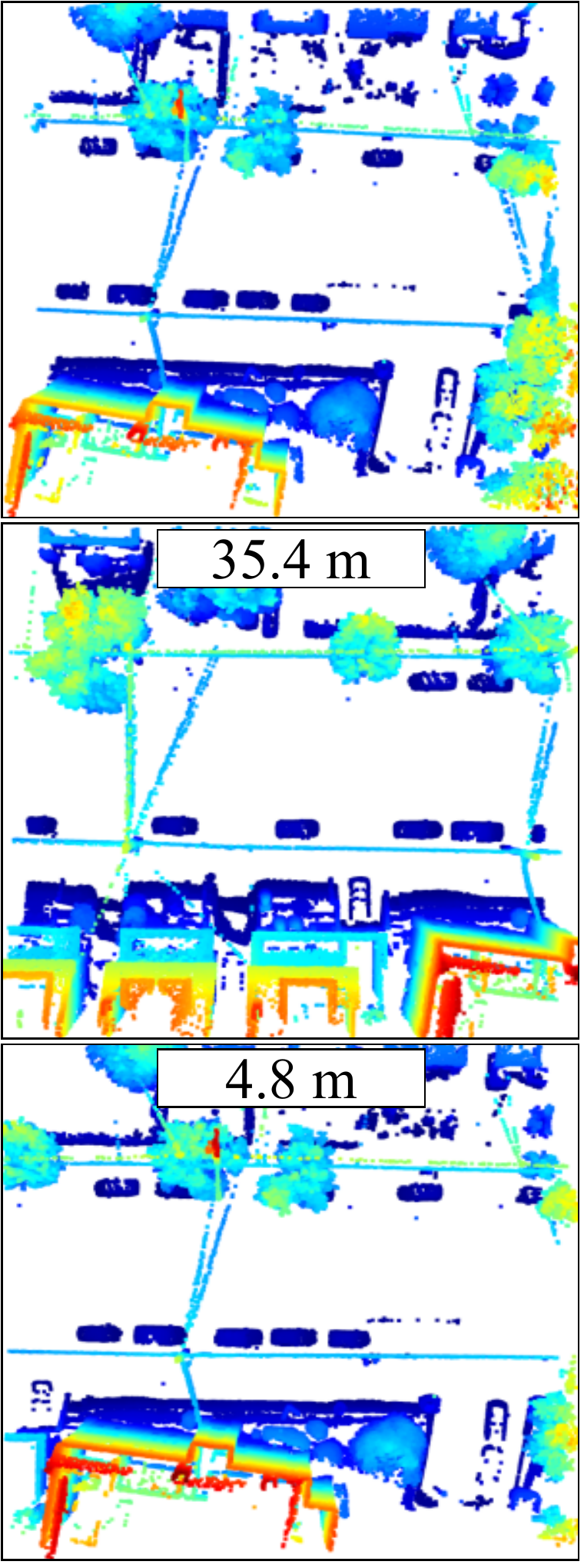}
         \caption{ALITA}
         \label{fig:qual_rr_alita}
     \end{subfigure}
     \hfill
     \begin{subfigure}[b]{0.1549\textwidth}
         \centering
         \includegraphics[width=\textwidth]{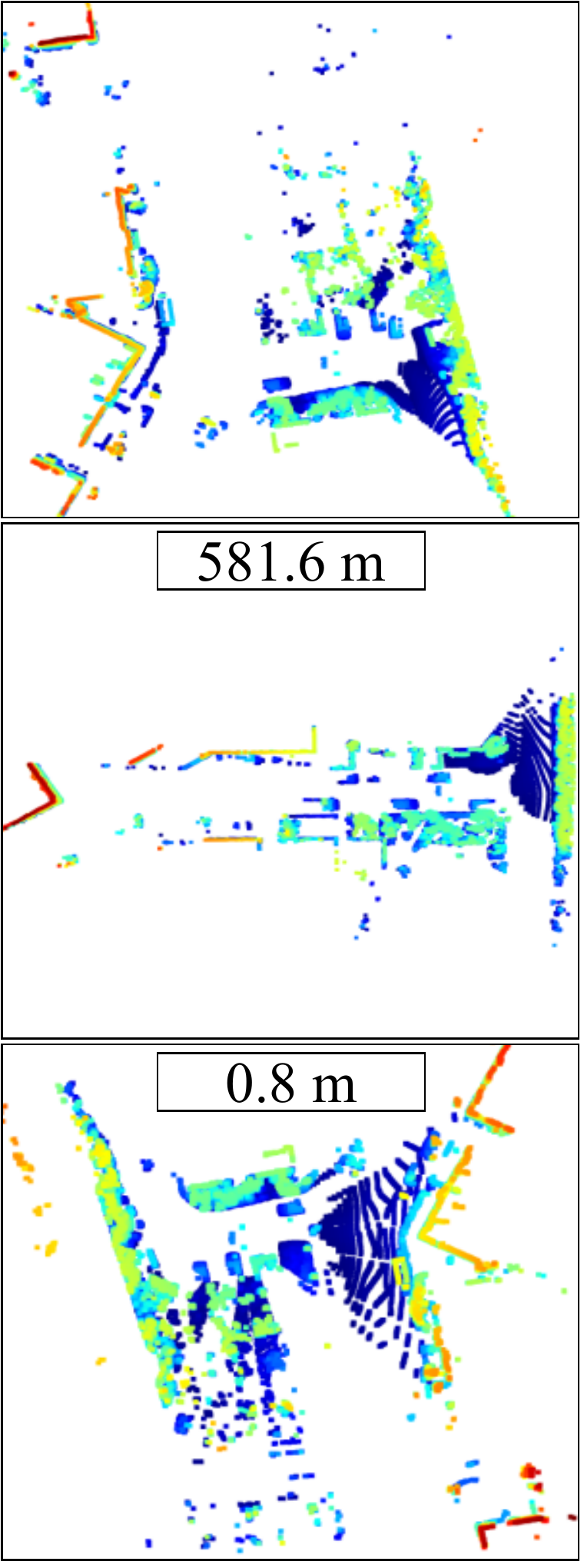}
         \caption{KITTI-360 09}
         \label{fig:qual_rr_kitti360}
     \end{subfigure}
    \caption{Visualizations of retrieval corrections by SpectralGV on the five datasets using the EgoNN architecture. Our re-ranking method is able to filter out incorrect initial candidates which are highly structurally similar to the query and correct them by accurately identifying the correct candidate. The distance of each retrieval to the query is noted in the text box. 
    }
    \label{fig:quali_rerank}
\end{figure*}

%% file: chapters/tables/table_placog.tex
\begin{table*}[t]

\begin{center}
\begin{tabular}{l@{\quad}|c@{\quad}c@{\quad}c@{\quad}c@{\quad}|c@{\quad}c@{\quad}c@{\quad}c@{\quad}|c@{\quad}c@{\quad}c@{\quad}c@{\quad}|c@{\quad}c@{\quad}c@{\quad}c@{\quad}}
&  \multicolumn{4}{c}{MulRan Sejong} & \multicolumn{4}{c}{Apollo-SouthBay} & \multicolumn{4}{c}{KITTI} & \multicolumn{4}{c}{Average}  \\
&  \multicolumn{2}{c}{5\,m } & \multicolumn{2}{c}{20\,m } 
&  \multicolumn{2}{c}{5\,m } & \multicolumn{2}{c}{20\,m } 
&  \multicolumn{2}{c}{5\,m } & \multicolumn{2}{c}{20\,m } 
&  \multicolumn{2}{c}{5\,m } & \multicolumn{2}{c}{20\,m } 
\\
& \begin{tabular}{@{}c@{}}R1\end{tabular}
& \begin{tabular}{@{}c@{}}R5\end{tabular}
& \begin{tabular}{@{}c@{}}R1\end{tabular}
& \begin{tabular}{@{}c@{}}R5\end{tabular}
& \begin{tabular}{@{}c@{}}R1\end{tabular}
& \begin{tabular}{@{}c@{}}R5\end{tabular}
& \begin{tabular}{@{}c@{}}R1\end{tabular}
& \begin{tabular}{@{}c@{}}R5\end{tabular}
& \begin{tabular}{@{}c@{}}R1\end{tabular}
& \begin{tabular}{@{}c@{}}R5\end{tabular}
& \begin{tabular}{@{}c@{}}R1\end{tabular}
& \begin{tabular}{@{}c@{}}R5\end{tabular}
& \begin{tabular}{@{}c@{}}R1\end{tabular}
& \begin{tabular}{@{}c@{}}R5\end{tabular}
& \begin{tabular}{@{}c@{}}R1\end{tabular}
& \begin{tabular}{@{}c@{}}R5\end{tabular}

\\[2pt]
\hline
MinkLoc3D*~\cite{Komorowski_2021_WACV} & 82.3 & 94.4 & 92.1 & 96.6 & 77.2 & 93.8 & 95.0 & 98.3 & 95.7 & 96.9 & 97.7 & 98.0 & 85.0 & 95.0 & 94.9 & 97.6 \\
DiSCO*~\cite{xu2021disco} & 94.0 & 97.5 & 95.8 & 98.1 & 
95.1 & 96.6 & 95.4 & 97.0 & 90.7 & 91.3 & 92.3 & 94.5  & 93.2 & 95.1 & 94.5 & 96.5 \\
DH3D*~\cite{du2020dh3d} & 32.4 & 56.2 & 58.9 & 75.5 & 25.3 & 49.6 & 50.4 & 70.7 &  75.5 & 91.1 & 86.8 & 96.1  & 44.4 & 65.6 & 65.3 & 80.7 \\
ScanContext~\cite{kim2018scan}  & 86.1 & 89.6 & 88.5 & 91.5 & 91.0 & 91.0 & 92.1 & 92.4 & 96.3 & 97.4 & 96.3 &  97.6  & 91.1 & 92.6 & 92.3 & 93.8 \\ 
Locus~\cite{vidanapathirana2020locus} & 	67.0 & 75.8 & 70.6 & 74.6 &  85.5 & 95.5 & 94.9 & 98.1 & 99.0 & 99.7 & 99.8 & \textbf{100}  & 83.8 & 90.3 & 88.4 & 90.9  \\
EgoNN~\cite{egonn} & 98.3 &  \textbf{99.9} &  \textbf{99.6} &  99.9 & 
95.7 & 97.7 & 96.3 & 98.2 & 
97.4 &  98.2 &  97.9 &  98.7
 & 97.1 & 98.6 & 97.9 & 98.9
 \\

LCDNet~\cite{cattaneo2022lcdnet} &  63.1 & 82.0 & 85.8 & 92.1 & 64.7 & 81.7 & 87.4 & 92.1  &  96.9 & 99.4 & 99.5 & 99.8  & 74.9 & 87.7 & 90.9 & 94.6  \\
LoGG3D-Net~\cite{vid2022logg3d} & 97.7 & 99.7 & 98.8 & 99.8 &  95.2 & 98.4 & 98.2 & \underline{99.2} &  \textbf{99.7} & \textbf{99.8} & \textbf{100} & \textbf{100}  & 97.5 & \underline{99.3} & 99.0 & \underline{99.6}  \\
\hline
EgoNN + SGV & \textbf{99.1} &  \textbf{99.9} &  \textbf{99.6} &  \textbf{100} & 
\underline{98.3} & \underline{98.7} & \underline{99.1} & \underline{99.2} & 
99.2 &  99.4 &  99.7 &  99.7  & \underline{98.8} & \underline{99.3} & \underline{99.4} & \underline{99.6}
 \\
LCDNet + SGV & 90.8 & 92.4 & 94.3 & 95.8 &  90.0 & 90.8 & 95.2 & 95.6 &  \textbf{99.7} & 99.7 & \textbf{100} & \textbf{100}   & 93.5 & 94.3 & 96.5 & 97.1 \\
LoGG3D-Net + SGV & \underline{99.0} & \textbf{99.9} & 99.5 & \textbf{100} &  \textbf{98.6} & \textbf{99.3} & \textbf{99.6} & \textbf{99.6} &  \textbf{99.7} & \textbf{99.8} & \textbf{100} & \textbf{100}  & \textbf{99.1} & \textbf{99.6} & \textbf{99.7} & \textbf{99.8} \\[2pt]
\end{tabular}
\caption{Place recognition on the `easy' evaluation set. 
(*) Results are obtained from \cite{egonn}. `R1': Recall@1, `R5': Recall@5.
}
\label{tab:place_recog_easy}
\end{center}
\end{table*}

%% file: chapters/tables/table_placog_hard.tex
\begin{table*}[t]

\begin{center}
\begin{tabular}{l@{\quad}|c@{\quad}c@{\quad}c@{\quad}c@{\quad}c@{\quad}|c@{\quad}c@{\quad}c@{\quad}c@{\quad}c@{\quad}|c@{\quad}c@{\quad}c@{\quad}c@{\quad}c@{\quad}}
&  \multicolumn{5}{c}{MulRan DCC} & \multicolumn{5}{c}{ALITA} & \multicolumn{5}{c}{KITTI-360}  \\
&  \multicolumn{3}{c}{5\,m } & \multicolumn{2}{c}{20\,m } 
&  \multicolumn{3}{c}{5\,m } & \multicolumn{2}{c}{20\,m } 
&  \multicolumn{3}{c}{5\,m } & \multicolumn{2}{c}{20\,m } 

\\
& \begin{tabular}{@{}c@{}}R1\end{tabular}
& \begin{tabular}{@{}c@{}}MRR\end{tabular}
& \begin{tabular}{@{}c@{}} $\Delta$MRR\end{tabular}
& \begin{tabular}{@{}c@{}}R1\end{tabular}
& \begin{tabular}{@{}c@{}}MRR\end{tabular}
& \begin{tabular}{@{}c@{}}R1\end{tabular}
& \begin{tabular}{@{}c@{}}MRR\end{tabular}
& \begin{tabular}{@{}c@{}}$\Delta$MRR\end{tabular}
& \begin{tabular}{@{}c@{}}R1\end{tabular}
& \begin{tabular}{@{}c@{}}MRR\end{tabular}
& \begin{tabular}{@{}c@{}}R1\end{tabular}
& \begin{tabular}{@{}c@{}}MRR\end{tabular}
& \begin{tabular}{@{}c@{}}$\Delta$MRR\end{tabular}
& \begin{tabular}{@{}c@{}}R1\end{tabular}
& \begin{tabular}{@{}c@{}}MRR\end{tabular}

\\[2pt]
\hline
ScanContext~\cite{kim2018scan}  & 	72.0 & 80.5 & - & 93.2 & 94.0  & 57.1 & 66.5 & - & 59.5 &  69.4  & 93.7 & 94.9 & - &  	95.2 & 	95.9   \\
Locus~\cite{vidanapathirana2020locus}  & 46.3 & 55.6 & - & 71.0 & 77.5  & 95.2 & 97.6 & - & \textbf{100} &  \textbf{100}  & 92.7 & 95.1 & - &  97.0 & 98.1   \\

EgoNN~\cite{egonn}  & 67.1 & 77.2 & - & 90.2 & 92.3  & 79.8 & 87.3 & - & 91.7 &  94.9  & 86.9 & 90.1 & - &  88.2 & 91.4   \\
LCDNet~\cite{cattaneo2022lcdnet} & 57.7 & 71.6 & - & 90.9 & 93.8 &  
90.5 & 94.6 & - & 96.4 & 97.8 &  
85.9 & 89.7 & - & 95.7 & 96.6   \\
LoGG3D-Net~\cite{vid2022logg3d} & 66.8 & 76.7 & - & 91.2 & 92.9 &  
15.5 & 28.0 & - & 26.2 & 40.0 &  
93.2 & 95.5 & - & 97.5 & 98.1   \\
\hline
 EgoNN + SGV & \underline{73.6} & \textbf{82.2} & 5.0 & 95.1 & 95.7  & \textbf{100} & \textbf{100} & 12.7 & \textbf{100} &  \textbf{100}  & \underline{95.7} & \underline{96.9} & 6.8 &  98.7 & 98.8   \\
LCDNet + SGV & 72.3 & 83.0 & 11.4 & \textbf{99.3} & \textbf{99.4} & 
\textbf{100} & \textbf{100} & 5.4 & \textbf{100} & \textbf{100} &  
\underline{95.7} & 96.2 & 6.5 & \underline{99.0} & \underline{99.1}   \\
LoGG3D-Net + SGV & \textbf{73.9} & \underline{82.0} & 5.3 & \underline{96.1} & \underline{96.5} &  
60.7 & 63.2 & 35.2 & 75.0 & 77.3 &  
\textbf{96.7} & \textbf{98.1} & 2.6 & \textbf{99.5} & \textbf{99.5} \\[2pt]
\end{tabular}
\caption{Place recognition on the unseen `hard' evaluation set. 
 $\Delta$MRR: increase in MMR after re-ranking in 5\,m threshold.  }
\label{tab:place_recog_hard}
\end{center}
\end{table*}

%% file: chapters/tables/table_metloc_rerank.tex
\begin{table}[t]

\begin{center}
\begin{tabular}{@{\enspace}l@{\enspace}|@{\enspace}c@{\enspace}|@{\enspace}c@{\enspace}|@{\enspace}c@{\enspace}}
\begin{tabular}{@{}c@{}}  \end{tabular}
& \begin{tabular}{@{}c@{}}S (\%) ($\uparrow$) \end{tabular}
& \begin{tabular}{@{}c@{}}RTE (m) ($\downarrow$) \end{tabular}
& \begin{tabular}{@{}c@{}}RRE ($^\circ$) ($\downarrow$) \end{tabular}
\\[1pt]
\hline

EgoNN~\cite{egonn}  & 88.2 & 42.45 & 13.36 \\
LCDNet~\cite{cattaneo2022lcdnet} & 94.0   & 10.40 & 6.37  \\
LoGG3D-Net~\cite{vid2022logg3d} &  97.0  & 9.51 & 3.64 \\
\hline
EgoNN + SGV  & 98.7  & 4.32 & 1.44 \\
LCDNet + SGV  & 98.5   & \textbf{0.79} & 2.35 \\
LoGG3D-Net + SGV  &  \textbf{99.5}  & 2.88 & \textbf{1.07} \\

\end{tabular}
\caption{Results of 6DoF metric localization on the unseen KITTI-360 09 dataset with and without re-ranking. 
}
\label{tab:met_loc_new}
\end{center}
\vspace{-6mm}
\end{table}

%% file: chapters/conclusion.tex
\section{Conclusion}
\label{sec:conclusion}

In this paper we explore re-ranking of point cloud retrievals for large-scale place recognition and metric localization.
We study multiple methods for re-ranking point cloud retrieval and introduce an efficient spectral method for geometric verification based re-ranking which does not require registering the query to each candidate. Our method can be readily integrated with current metric localization methods without any changes to their architecture. We show consistent and significant performance improvements in 5 different large-scale datasets in all evaluation settings. We empirically demonstrate how, in the context of point cloud retrieval, all geometric verification methods are robust, and Spectral Geometric Verification obtains the best performance improvement after re-ranking while being the only method efficient for real-time robotic application.